\documentclass{article}

\usepackage{arxiv}

\usepackage[utf8]{inputenc} 
\usepackage[T1]{fontenc}    
\usepackage{hyperref}       
\usepackage{url}            
\usepackage{booktabs}       
\usepackage{amsfonts}       
\usepackage{nicefrac}       
\usepackage{microtype}      
\usepackage{lipsum,subcaption}
\usepackage{IEEEtrantools}
\usepackage{graphicx}
\usepackage{epstopdf}
\newcommand{\ens}{\IEEEeqnarraynumspace}

\title{Performance comparison of 3D correspondence grouping algorithm for 3D plant 
point clouds}

\author{
  Shiva Azimi\thanks{Corresponding author.} \\
  Department of Electrical Engineering\\
  Indian institute of Technology Delhi\\
  New Delhi, India 110016 \\
  \texttt{shiva.azimi@yahoo.com} \\
   \And
 Tapan K. Gandhi \\
  Department of Electrical Engineering\\
  Indian institute of Technology Delhi\\
  New Delhi, India 110016 \\
  \texttt{tgandhi@ee.iitd.ac.in} \\
}

\begin{document}
\maketitle

\begin{abstract}
Plant Phenomics can be used to 
monitor the health and the growth of plants. Computer vision applications like stereo reconstruction, image retrieval, object tracking, and object recognition play an important role in imaging based plant phenotyping. This paper offers a comparative evaluation of some popular 3D correspondence grouping algorithms, motivated by the important role that they can play in tasks such as model creation, plant recognition and identifying plant parts. Another contribution of this paper is the extension of 2D maximum likelihood matching to 
3D Maximum Likelihood Estimation Sample Consensus (MLEASAC). MLESAC is efficient and is computationally less intense than 3D random sample consensus (RANSAC). We test these algorithms on 3D point clouds of plants along with two standard benchmarks addressing shape retrieval and point cloud registration scenarios. The performance is evaluated in terms of precision and recall.
\end{abstract}


\section{Introduction}
Providing food to increasing world population has become a challenging task due to loss of arable land, climatic changes and many man made factors. This requirement for increasing the crop yield has necessitated the development of modern tools and techniques for increasing crop yield \cite{challinor2014meta}. 
Plant phenotyping deals with the measurement of changes in phenomes in reaction to genetic and environmental changes and is an important plant science technique that can be used to monitor plant health and growth. 

Recently, image-based high-throughput plant phenotyping has become an important computer vision research problem \cite{fahlgren2015lights}. Measurements of visual plant features can be done using either two dimensional (2D)  \cite{minervini2015image}, \cite{panwar2014imaging} or three dimensional (3D) models \cite{liu2017novel} of the plant.
However, since 3D models can capture much more information regarding the plant physiology as compared to 2D models, 3D models are preferred for plant phenotyping.

Finding correct correspondences between 3D shapes is a cornerstone in 3D computer vision tasks like 3D object recognition \cite{buch2016local}, point cloud registration \cite{lei2017fast}, and 3D object categorization \cite{xiang2015data}. The first stage in local feature-based matching is detecting the keypoints on the surface and generating descriptors for every keypoints. Generating the vector feature descriptors surrounding each keypoint is computationally demanding, and thus, detecting keypoints is required to reduce the amount of computation required.  In the second stage, crude initial matches are generated for marking the similarities between two 3D shapes. However, the initial
matches have a high number of false positives due to reasons such as the residual errors from the proceeding steps like errors in keypoint localization, noise, point density variations, occlusion, overlaps, etc.
To ensure accurate transformation estimation and hypothesis formulation, outliers must be filtered out from the initial matching, making correspondence grouping (CG) important \cite{glent2014search}.

CG can play an important role in 3D model based plant phenotyping and automation of the farming process. Certain applications that require correspondence matching include creating 3D models, plant species recognition, recognition of plant parts, unsupervised template learning for fine-grained plant structures , etc. In autonomous farming also, techniques based on correspondence matching can be used in many ways. For instance, a general model of a harvesting machine can be produced in large numbers and then any farmer having no robotics skills can teach it on-site to identify and pick the required kind of fruit. In the same way, an autonomous weeding machine can be mass produced,
then taught to distinguish weeds from the crops at different kinds of farms of widely, improving farming efficiency.

Although, there are studies in the existing literature comparing the performance of 3D CG algorithms \cite{yang2017performance}, to the best of our knowledge, there is no work in the existing literature that deals specifically with the performance evaluation of the CG algorithms as applied to 3D point cloud data for plant health classification and plant growth analysis. Working with 3D point cloud data in the case of plants offers special challenges:
\begin{itemize}
\item
Plants are made up of very fine structures that makes it very difficult to make a perfect scan. Thus the keypoint detectors 
local feature descriptors have to be robust to noise and holes present in the point cloud data.
\item
The lighting conditions in farming keep changing, offering different illumination conditions. This also changes the color and the  texture of the plants, thus, making it very difficult to use the color information for decision making.
\end{itemize}

\begin{figure*}[h!]
    \centering
        \includegraphics[height=5.0cm, width=15cm]{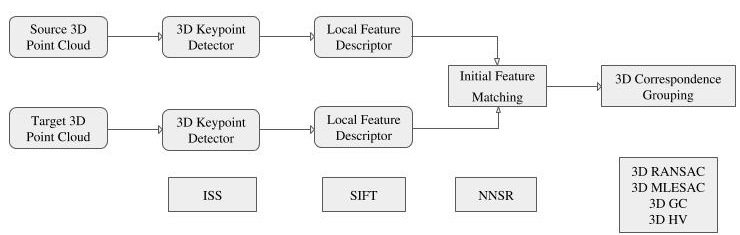}
        \label{figurelabel}
    \caption{Performance comparison of 3D correspondence grouping algorithms}\label{fig-plot}
\end{figure*}

In summary, in this paper, we consider four most common 3D CG algorithms, viz, Nearest Neighbor Similarity Ratio (NNSR) \cite{muja2012fast}, \cite{hietanen2017robustifying}, Random Sample Consensus (RANSAC) \cite{fischler1981random}, \cite{glent2014search}, Geometric Consistency (GC) \cite{chen20073d}, \cite{yang2019ranking} and 3D Hough Voting (3DHV) \cite{tombari2010object}, \cite{buch2017rotational}. We have also implemented 3D MLEASAC, which combines Maximum Likelihood Estimation (MLE) and 3D RANSAC and is motivated by the approach taken in \cite{torr2000mlesac}, \cite{deng2018research} for 2D images. Fig.\ref{fig-plot} is represented the pipeline of our work.

The remainder of the paper is structured as follows. Section II briefly overviews the related work. We give a description of the 3DCG algorithms that we have considered for evaluation in Section III. performance evaluation is provided in Section IV. Finally, in Section V, we provide some perspectives and conclude the paper.
\section{Related work}
3DCG is still an active area in computer vision. 3DCG works based on local shape feature matching and is commonly used in other applications such as object recognition, pose estimation, point cloud registration, etc.
Image based correspondences features have been explored in depth in the existing literature.  
Torresani \cite{torresani2008feature} presented a method for finding correspondences between parse image features by graph matching. They solve the matching problem by minimizing a image matching objective function which depend on the appearance, geometric and spatial features. This method for real-world image matching is able to achieve global optimal by an unknown non-rigid mapping.
Lowe \cite{lowe2004distinctive} is described how to use SIFT features for object recognition in image. In this paper, they investigated Nearest Neighborhood algorithm for matching individual feature to a database of feature base on Euclidean distance of their feature vector. They found good matches in a large number of features with using distinctive descriptors. This method is extended to 3D in \cite{glent2014search}. 
In \cite{papazov2010efficient} is modified RANSAC method for 3D object recognition and pose estimation. They obtained optimum number of iteration by using the combination algorithms of descriptor, hash table and RANSAC. 
In \cite{chen20073d}, Chen et al. used geometric constraints for 3DCG algorithm in 3D object recognition. They improved the version of Johnson et al. \cite{johnson1998surface} algorithm for clustering the inliers into disjointed clusters. 
Tombari et al. \cite{tombari2010object} introduced 3D Hough Voting for 3D object recognition by using Hough Transform \cite{hough1962method}. In the proposed approach each correspondence can cast a vote for presence of the unique reference point of object in the 3D Hough space simultaneously.
Anders et al. \cite{glent2014search} introduced two stage local and global method for casting a vote for finding 3DCG. On the local stage, they caste a vote for finding correspondences by geometric consistency and in the global stage, they cast a vote by covariance constraint. At the final stage, they compared their algorithm with existing ones.
\section{3D correspondence grouping algorithms }
In this section, we provide a brief overview the 3DCG algorithms. Given a model $\mathcal{S}$ and a scene $\mathcal{S'}$, where an raw correspondence set $\mathcal{C}$ is obtained after comparing the feature sets $\mathcal{F}$ and $\mathcal{F'}$ belongng to $\mathcal{S}$ and $\mathcal{S'}$, respectively, the goal is to obtain a subset $\mathcal{C}_{inlier}$ of $\mathcal{C}$ that identifies the correct correspondences between $\mathcal{S}$ and $\mathcal{S'}$, i.e., the inliers. 
An element in $\mathcal{C}$ is given as:
$c=\{p, p', s_{\mathcal{F}}\left(f, f' \right)\}$, where $p$ and $p'$ are keypoints belonging to the model (source) and the scene (target), respectively, $f$ and $f'$ are the descriptors for these keypoints, and $s_{\mathcal{F}}\left(f, f' \right)$ is the similarity score assigned to $c$. We now briefly explain the 3DCG
algorithms considered in this paper.
\subsection{Nearest neighbor similarity ratio}
NNSR is a basic method for determining the initial correspondences \cite{muja2012fast}. This method is based on Lowe's ratio \cite{lowe2004distinctive}. It judges match features base on the ratio of two closest neighbors and obtained by Euclidean ($L2$) distance between an invariant feature vector $f \in \mathcal{F}$ with the nearest  $f'_{1}\in \mathcal{F'}$ and second nearest matching feature $f'_{2} \in \mathcal{F'}$. A correspondence will be as an inlier if the Euclidean distance between $f$, $f'_{1}$ and  $f$, $f'_{2}$  satisfies: 
\begin{IEEEeqnarray}{C}
\frac{\left\|{f - f'_{1}} \right\|_{L2}}{\left\|{f - f'_{2}} \right\|_{L2}} \le {t_{nnsr}},
\label{eq1}
\end{IEEEeqnarray}
where, $t_{nnsr}$ is the threshold. The NNSR will be used as a baseline for all comparison purposes.
\subsection{RANSAC}
RANSAC is an iterative method for judging the correctness of a correspondence set by calculating the number of inliers returned in each iteration \cite{fischler1981random}. The RANSAC method uses random samples from the initial correspondences set $\mathcal{C}$ in each iteration and finds the transformation $T$ corresponding to those samples. The steps are repeated $N_{ransac}$ iteration and the transformation is compared with a threshold $t_{ransac}$ . The transformation returning the maximum number of inliers is judged the optimal $T^*$. The correspondences in $\mathcal{C}$ that do not agree with $T^*$ are classified as outliers.
 
\subsection{Geometric consistency}
CG method is based on geometric constrains  for  finding  inliers and is not dependent on the  local feature space \cite{chen20073d}. The  algorithm looks for consistency and compatibility in terms of consistent cluster. After  finding initial matching set $\mathcal{C}$, we find the distance between keypoints from  the  seed  in  the model and scene and  then  we  find $L2$-distances between the pairwise correspondences to compare with the geometric constraint threshold $t_{gc}$. Finally, we choose the largest cluster as optimum one and classify the corresponding correspondences as inliers. 
The compatibility score for the given correspondences pair $c_1$ and $c_2$ is given as:
\begin{IEEEeqnarray}{C}
d\left( {c_{1},c_{2}} \right) = \left| {d\left( {p_{1},p_{2}} \right) - d\left( {p_{1}',p_{2}'} \right)} \right| \le {t_{gc}},
\label{eq2}
\end{IEEEeqnarray}
where ${d\left(p_{1},p_{2}\right)=\left\|{(p_{1}-p_{2}}|\right|_{L2}}$.

\subsection{3D Hough voting} 
3DHV uses features by casting a vote in 3D Hough space for finding inliers \cite{tombari2010object}. In this method, after finding initial matching set $\mathcal{C}$, for each keypoint $p_i$ in the source shape (model) that is in $\mathcal{C}$ , the vector between source shape centeroid $C_S$ and $p_i$ is computed in the global reference frame (GRF):
\begin{IEEEeqnarray}{C}
{\bf{V}}_{i,G}^S = {C_S} -p_{i},
\label{eq3}
\end{IEEEeqnarray}
After that $ {\bf{V}}_{i,G}^S $ is transformed to local reference frame (LRF), as we want our method to be translation and rotation invariant.
\begin{IEEEeqnarray}{C}
{\bf{V}}_{i,L}^S = {\bf{R}}_i^S \cdot {\bf{V}}_{i,G}^S,
\label{eq4}
\end{IEEEeqnarray}
Where ${\bf{R}}_i^S$ is rotation matrix, and each line in ${\bf{R}}_i^S$ being a unit
vector of the LRF of $p_i$. Similarly, we can obtain ${\bf{V}}_{i,G}^{S'}$ corresponding to $p_i'$. Then, ${{\bf{V}}_{i,L}^S=\bf{V}}_{i,L}^{S'}$, because of the  invariance in LRF. The vector ${\bf{V}}_{i,G}^{S'}$ is computed by transforming ${\bf{V}}_{i,L}^{S'}$ into GRF. \begin{IEEEeqnarray}{C}
{\bf{V}}_{i,G}^{S'} = {\bf{R}}_i^{S'}\cdot{\bf{V}}_{i,L}^{S'} + {p_i}',
\label{eq5}
\end{IEEEeqnarray}
Eventually the feature ${f_i}'$ could cast a vote in 3D Hough space through the vector 
${\bf{V}}_{i,G}^{S'}$. The inliers correspond to the peak of Hough space.
\subsection{3D MLESAC}  
MLEASAC combines 3D RANSAC with MLE for finding inliers. The novelty of our algorithm is generalization of RANSAC in 3D. This algorithm modifies RANSAC in a way such that we look for the transformation that maximizes the likelihood function
rather than maximizing the number of inliers. Our implementation of 3D MLESAC consists of the following basic steps.

Similar to 3D RANSAC, given $N_{mlesac}$ iterations, in each iteration, five random components are sampled from the initial matching $\mathcal{C}$.
We use these samples to calculate the affine transformation $T$ and we transform all model keypoint that are in $\mathcal{C}$ according to $\mathcal{C}$. We then define the likelihood function for each $i=1,...,N$, with $N$ being the number of column entries in $\mathcal{C}$  as:
\begin{IEEEeqnarray}{C}
\Pr \left( d_i \right) = \gamma \frac{1}{{\sqrt {2\pi {\sigma ^2}} }}\exp \left( { - \frac{{{d_i^2}}}{{2{\sigma ^2}}}} \right) + \left( {1 - \gamma } \right)\frac{1}{\nu },
\label{eq6}
\end{IEEEeqnarray}
where, $d_i$ is the Euclidean distance between transformed model keypoints $p_i$ with matched points $p_i'$ in scene, $\nu$ is a constant, $\sigma$ is the standard deviation, and $\gamma$ is the mixing parameter, which has been calculated according to the method described in \cite{torr2000mlesac}. 

Since, distances $d_i$ between $p_{i}^S$ and $p_{i'}^{S'}$ are independent of each other, the joint likelihood function for all $d_is$ given as:
\begin{IEEEeqnarray}{C}
\Pr \left( d \right) = \prod\limits_{i = 1}^N {\left( {\gamma \frac{1}{{\sqrt {2\pi {\sigma ^2}} }}\exp \left( { - \frac{{{d_i}^2}}{{2{\sigma ^2}}}} \right) + \left( {1 - \gamma } \right)\frac{1}{\nu }} \right)} .\ens
\label{eq7}
\end{IEEEeqnarray}
To simplify the calculations, we take logarithm on both sides of the above, giving the negative log-likelihood function as;
\begin{IEEEeqnarray}{c}
 - L =  - \sum\limits_{i = 1}^N {\log \left( {\gamma \frac{1}{{\sqrt {2\pi {\sigma ^2}} }}\exp \left( { - \frac{{{d_i}^2}}{{2{\sigma ^2}}}} \right) + \left( {1 - \gamma } \right)\frac{1}{\nu }} \right)}  .\ens
\label{eq8}
\end{IEEEeqnarray}

Finally, the transformation that results in the minimum of negative log-likelihood function is compared with $t_{mlesac}$  and it is considered the optimum transformation $T^*$. The correspondences in $\mathcal{C}$ that do not agree with $T^*$ are classified as outliers.

Note that the first term on the RHS of (\ref{eq6}), refers to the correspondences that are inliers while the second term corresponds to the outliers, and is assumed to have a uniform distribution.
\begin{figure}[ht!]
    \centering
    \begin{subfigure}[b]{0.30\textwidth}
        \includegraphics[height=3.0cm, width=3.5cm]{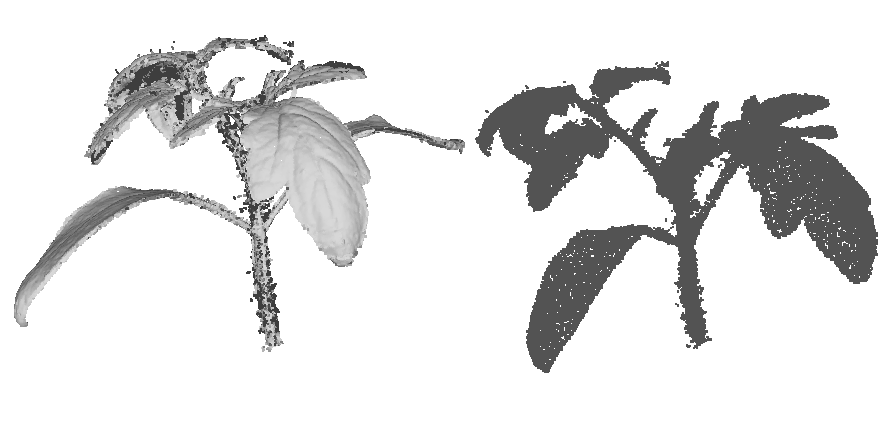}
        \caption{Plant}
        \label{fig}
    \end{subfigure}
    \begin{subfigure}[b]{0.30\textwidth}
        \includegraphics[height=3.0cm, width=3.5cm]{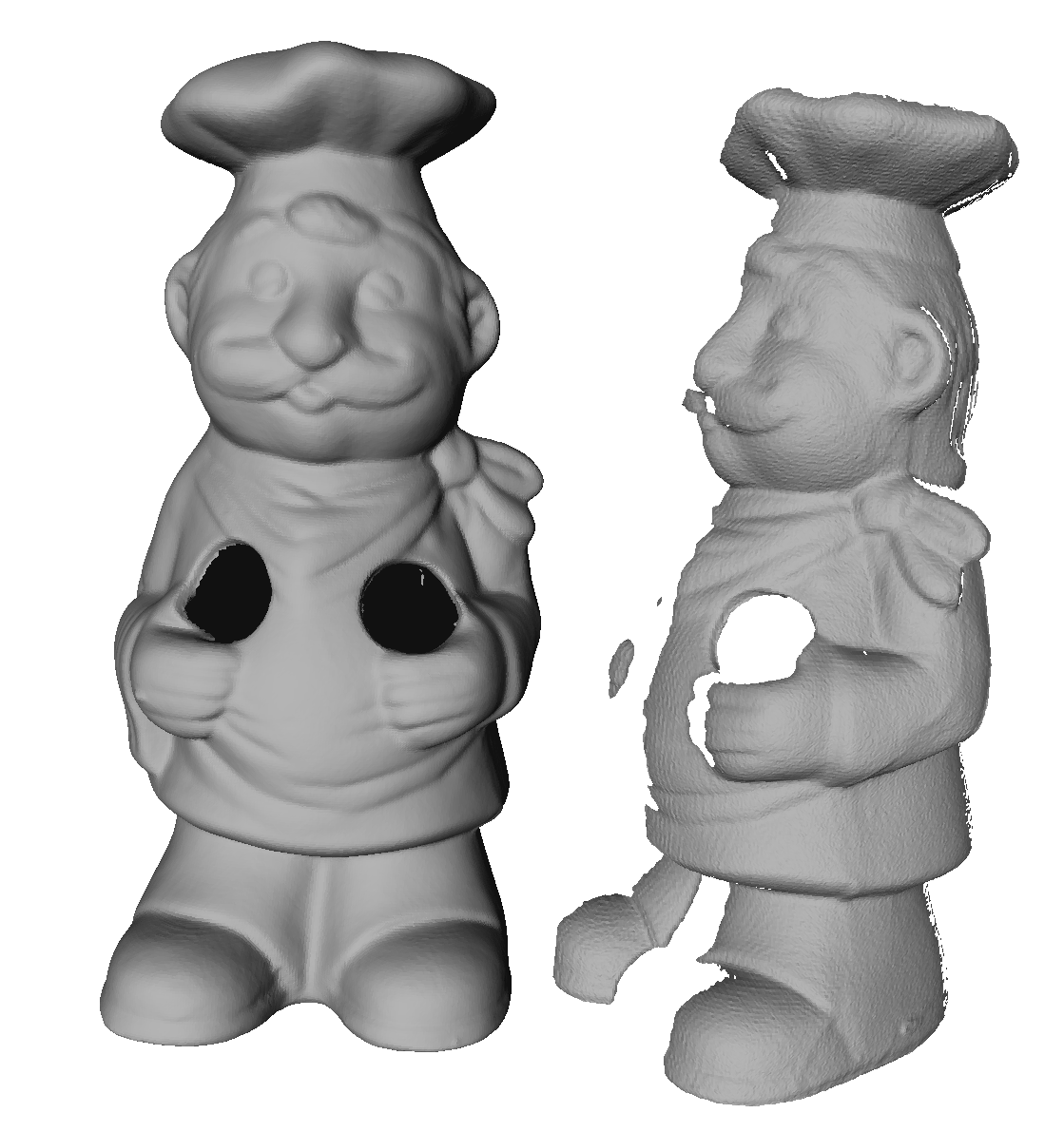}
        \caption{B3R}
        \label{fig:mouse}
    \end{subfigure}
    \begin{subfigure}[b]{0.30\textwidth}
        \includegraphics[height=3.0cm, width=3.5cm]{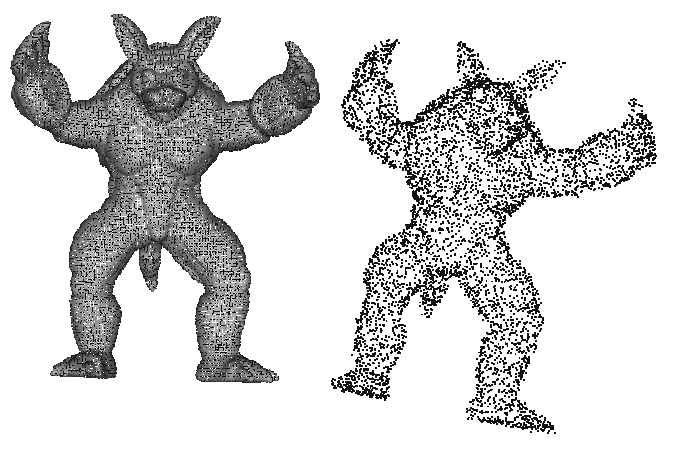}
       \caption{U3M}
        \label{fig:mouse}
    \end{subfigure}
    \caption{Visualization (snapshot) of sample model (right) and scene (left) from each dataset}\label{fig-b}
\end{figure}
\section{Experimental evaluation and discussion}
\begin{figure*}[ht!]
    \centering
    \begin{subfigure}[b]{0.45\textwidth}
       \includegraphics[height=4.5cm, width=6.5cm]{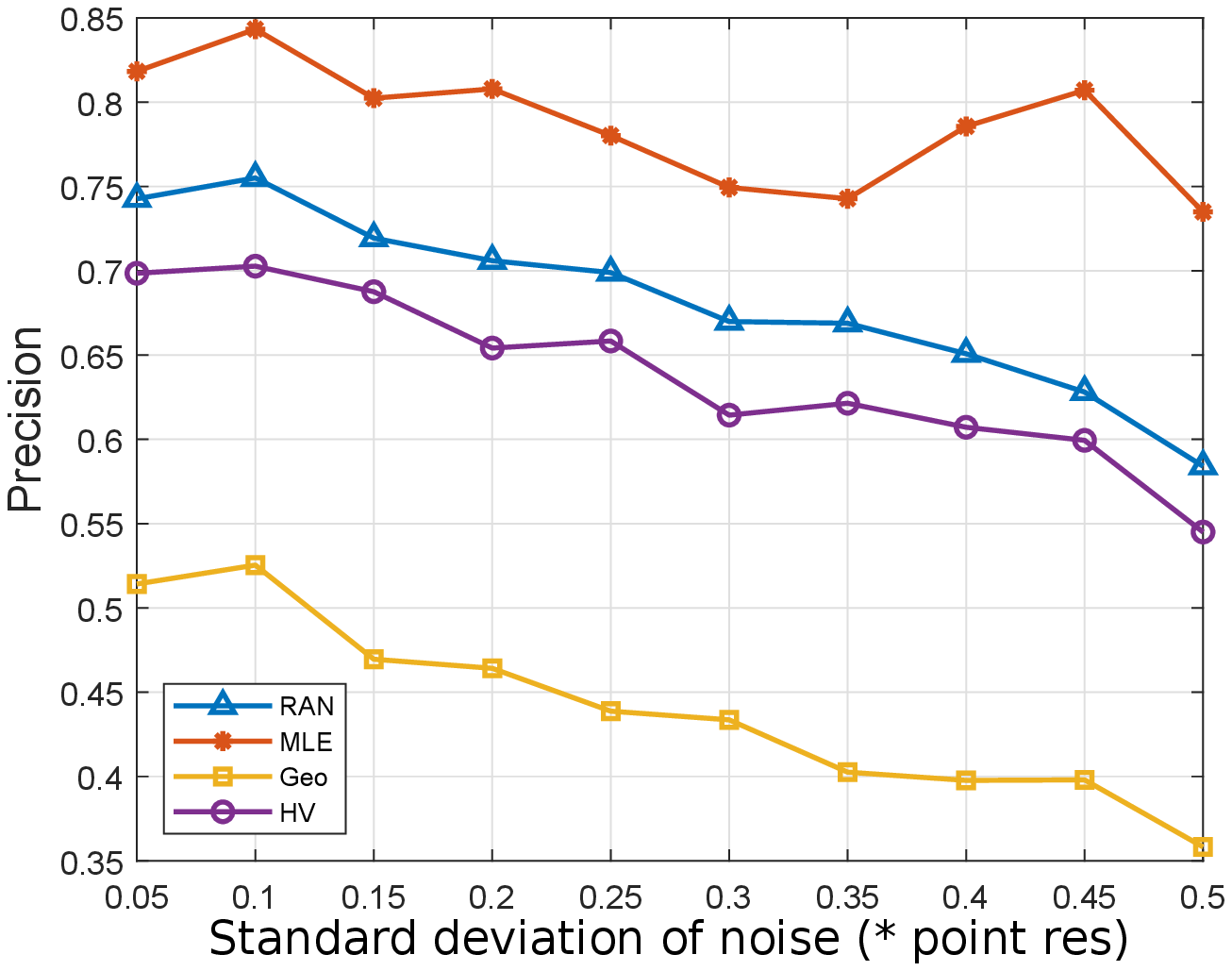}
    \end{subfigure}
    \begin{subfigure}[b]{0.45\textwidth}
        \includegraphics[height=4.5cm, width=6.5cm]{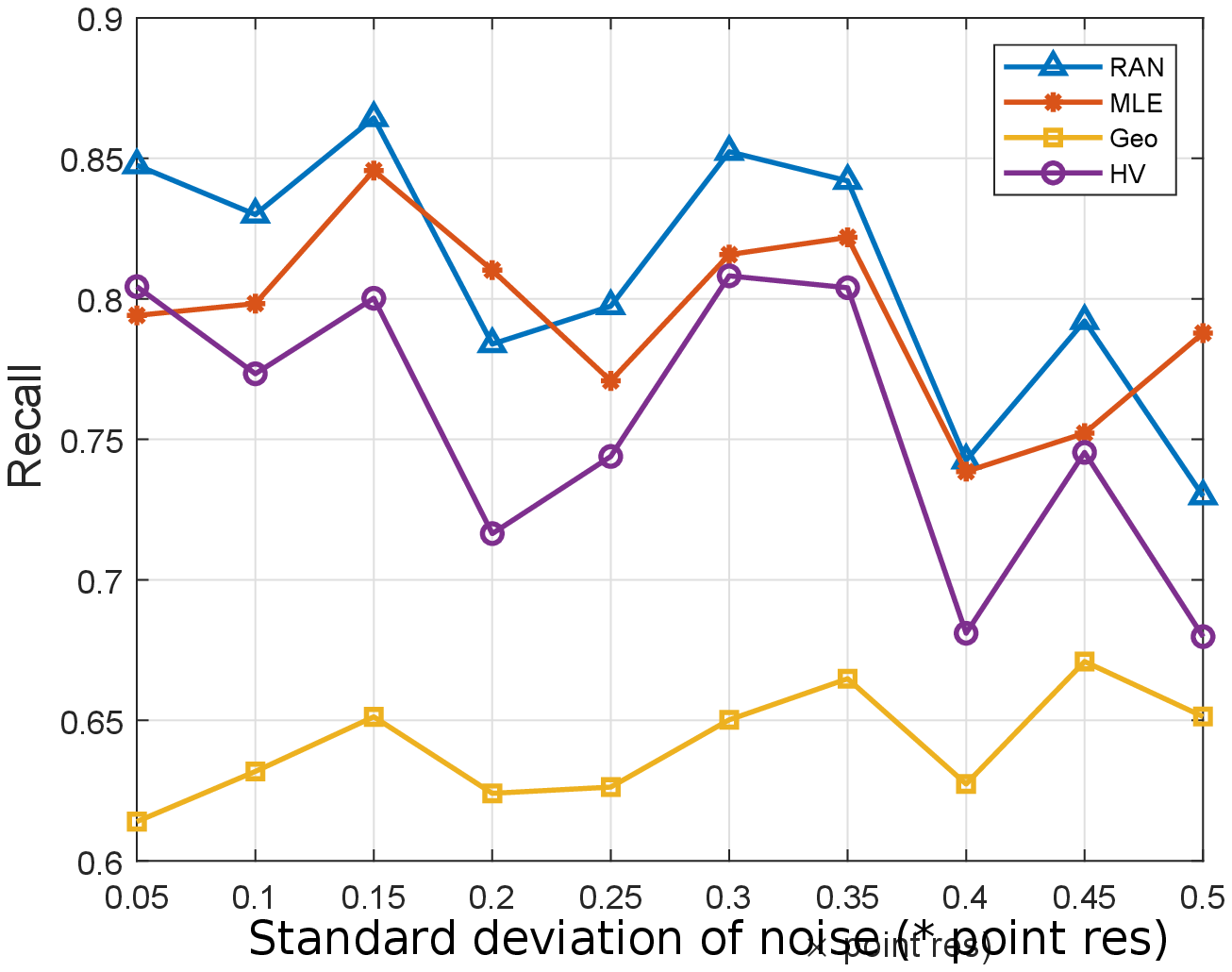}
   \end{subfigure}
    \begin{subfigure}[b]{0.45\textwidth}
        \includegraphics[height=4.5cm, width=6.5cm]{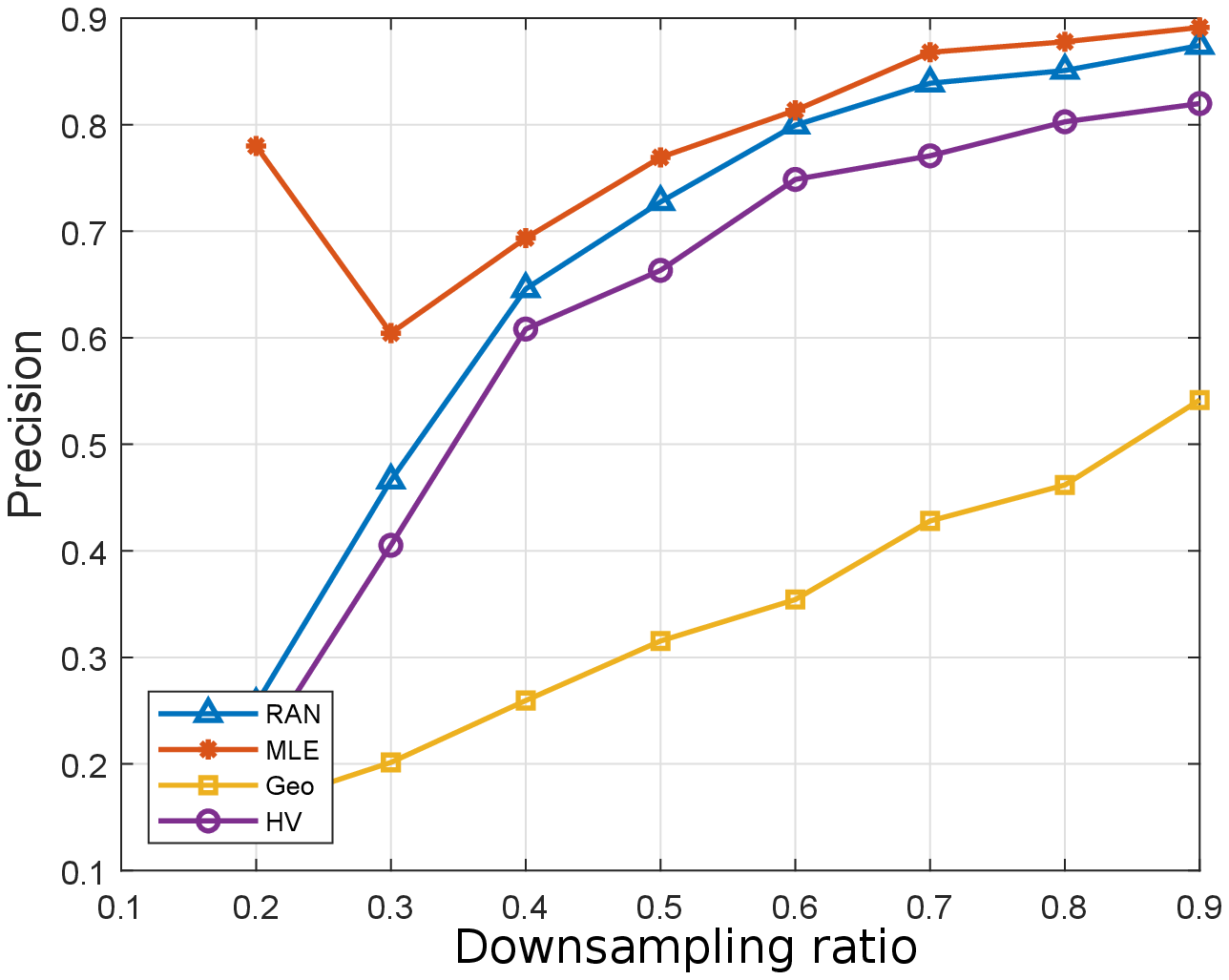}
    \end{subfigure}
   \begin{subfigure}[b]{0.45\textwidth}
        \includegraphics[height=4.5cm, width=6.5cm]{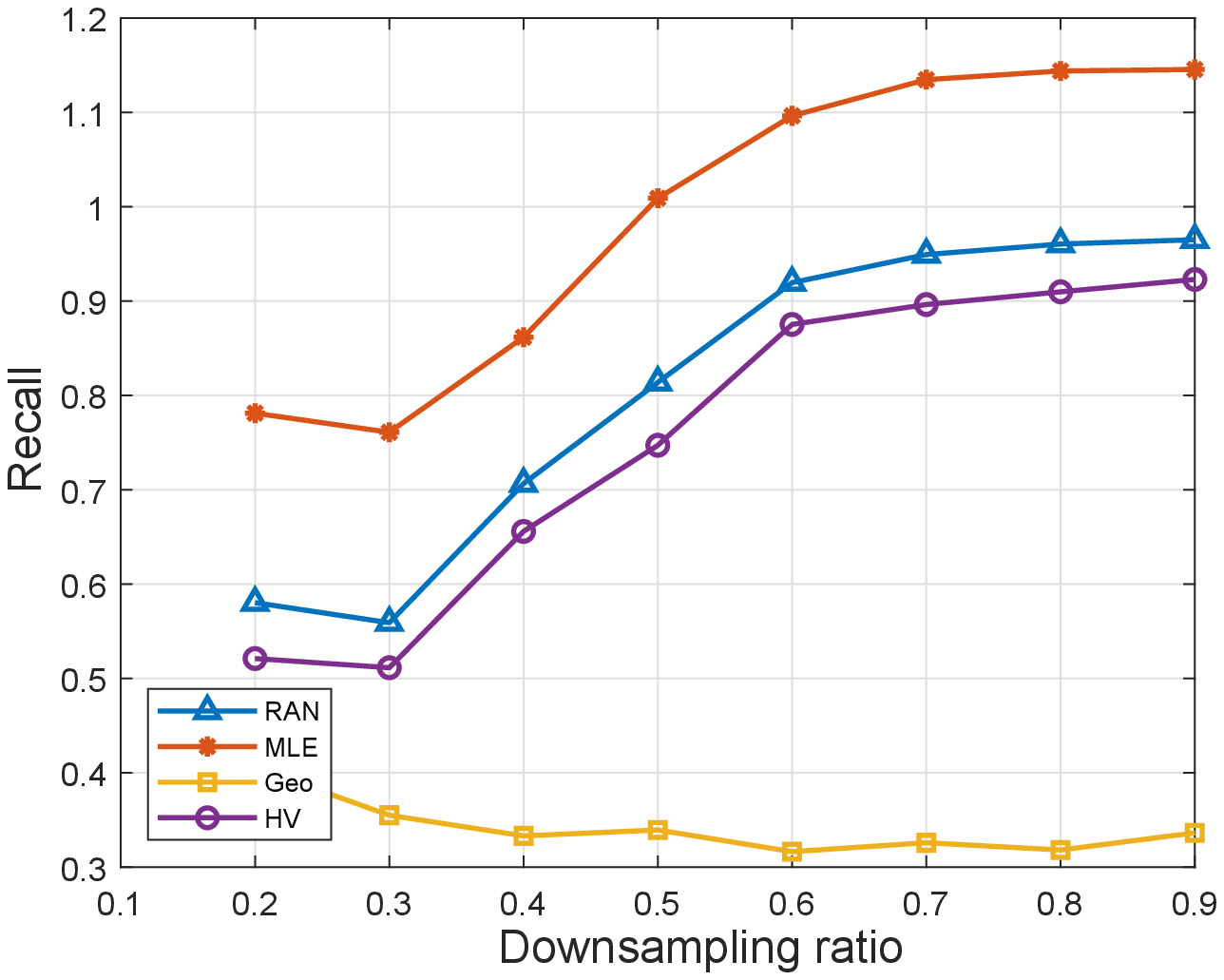}
    \end{subfigure}
    \caption{Performance of the evaluated CG algorithms in terms of
    precision and recall for the plant dataset}\label{fig-plant}
\end{figure*}  
\begin{figure*}[ht!]
    \centering
    \begin{subfigure}[b]{0.48\textwidth}
        \includegraphics[height=4.5cm, width=6.5cm]{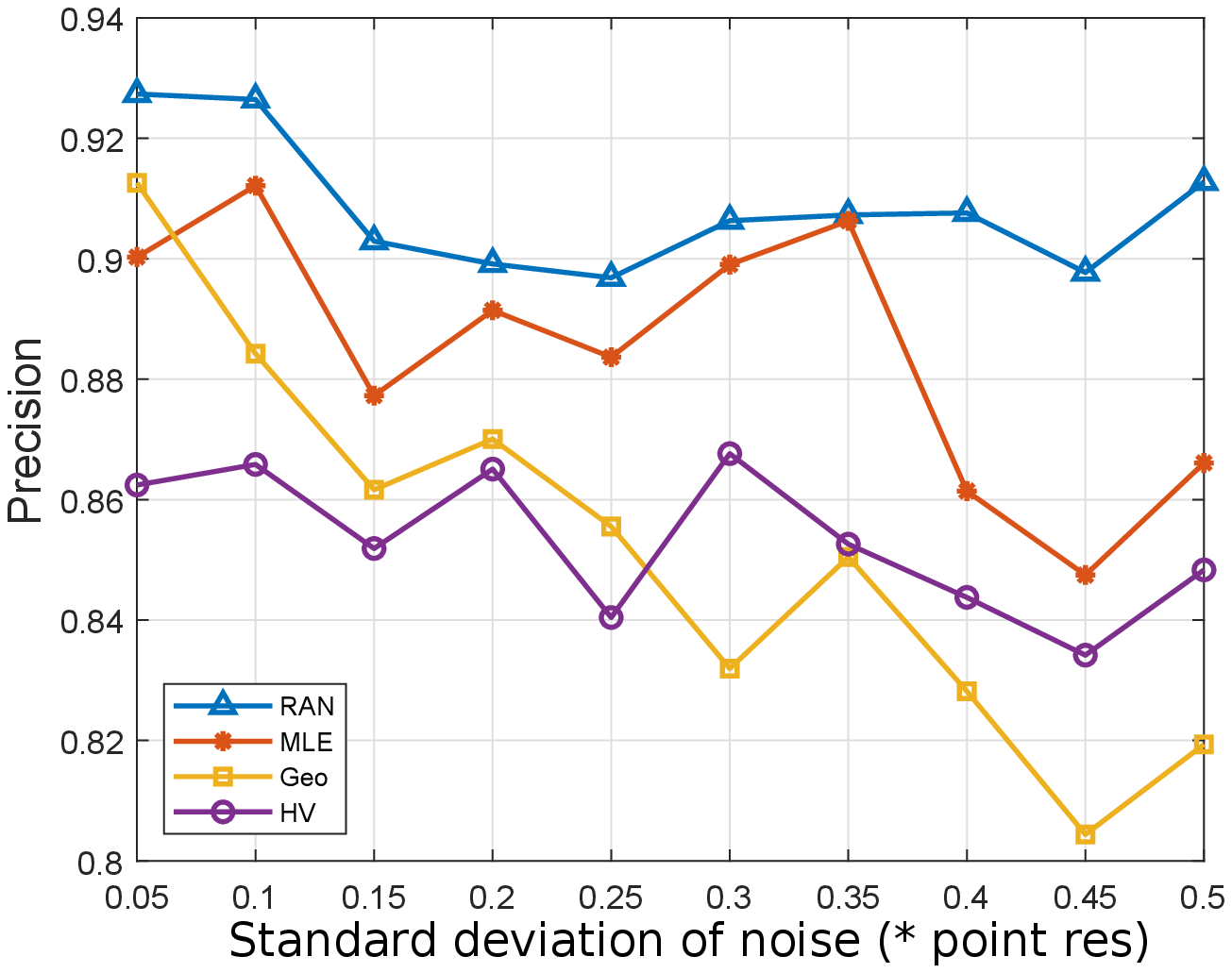}
        \label{fig:gull}
    \end{subfigure}
    \begin{subfigure}[b]{0.48\textwidth}
        \includegraphics[height=4.5cm, width=6.5cm]{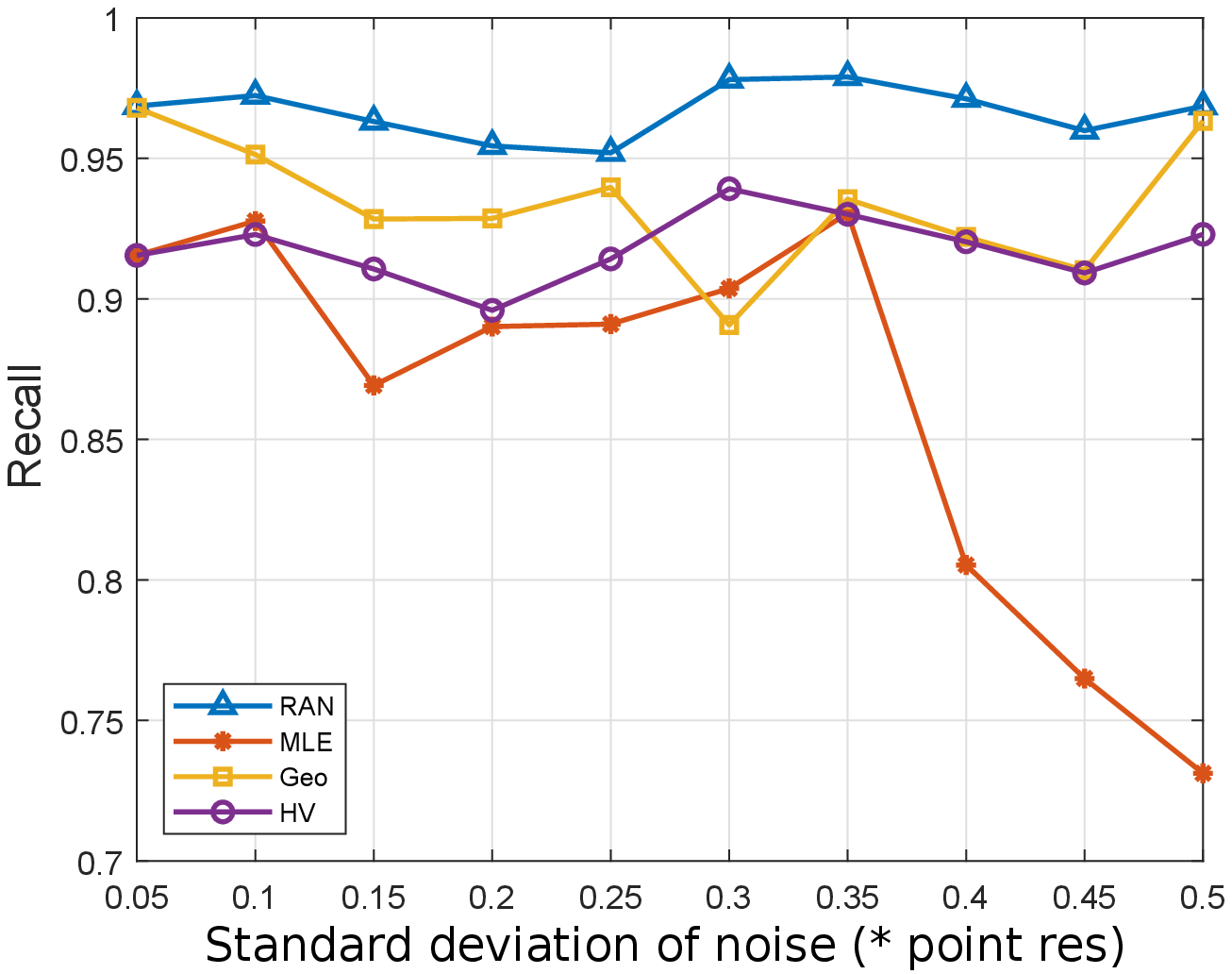}
        \label{fig:tiger}
    \end{subfigure}
    \begin{subfigure}[b]{0.48\textwidth}
        \includegraphics[height=4.5cm, width=6.5cm]{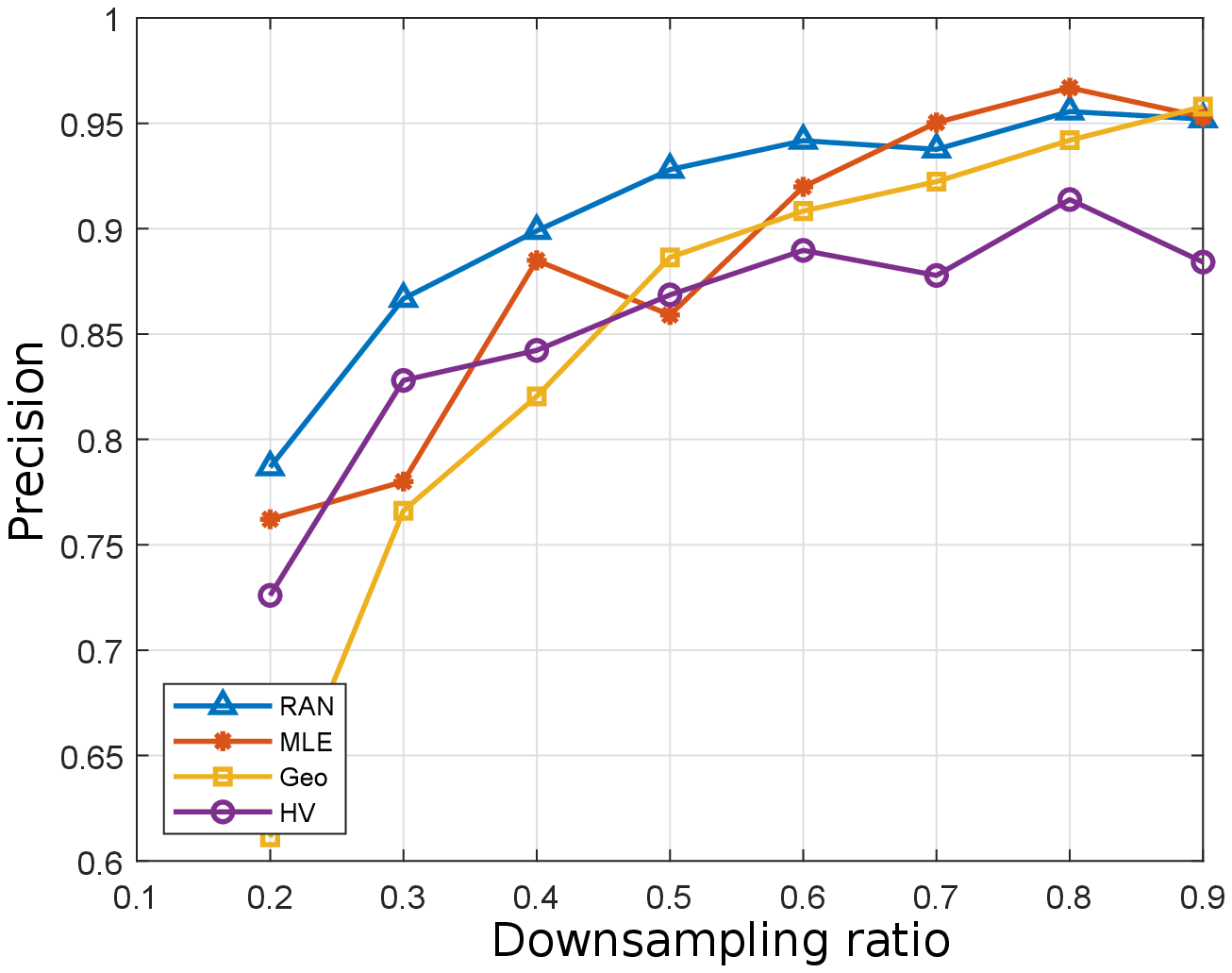}
        \label{fig:mouse}
    \end{subfigure}
    \begin{subfigure}[b]{0.48\textwidth}
        \includegraphics[height=4.5cm, width=6.5cm]{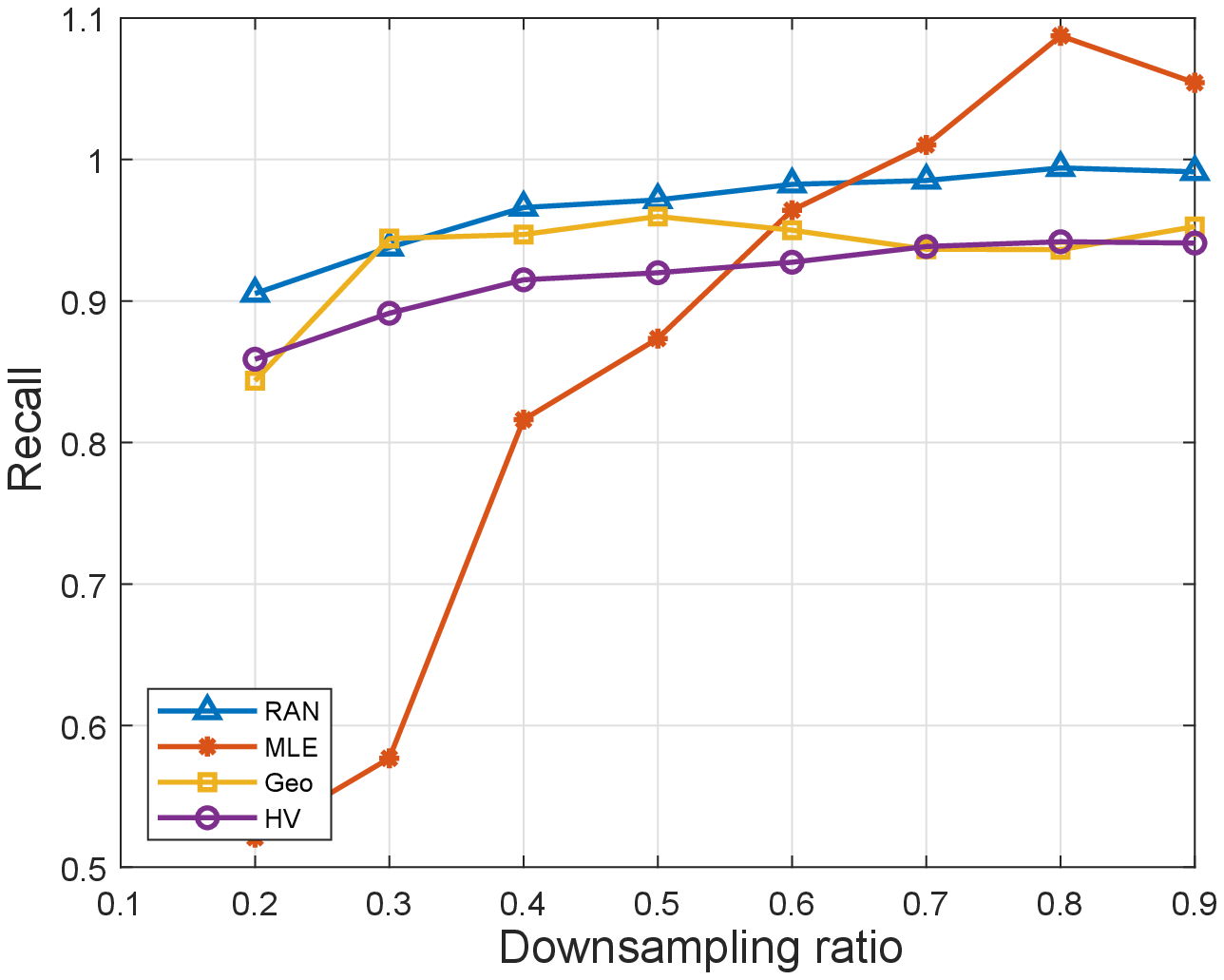}
        \label{fig:mouse}
    \end{subfigure}
    \caption{Performance of the evaluated CG algorithms in terms of
    precision and recall for the B3R dataset}\label{fig-b3r}
\end{figure*}  
In this section, we evaluate and provide the implementation details of all the algorithms discussed in the previous section on two chosen benchmarks and a plant dataset. Fig.\ref{fig-b} shows the samples of models and scenes in different datasets. These datasets have scenes with different noise levels, point densities, and partial
overlaps. We determine the correctness of a correspondence using the ground truth affine transformation $\textbf{T}_{GT}$ between the model and the scene. A given correspondence $c$ between points $p$, $p'$ is declared correct if:
\begin{IEEEeqnarray}{C}
\left| {{{\bf{T}}_{GT}}\left( p \right) - p'} \right| \le t,
\end{IEEEeqnarray}
Where $t$ is the threshold. The calculated inliers by all algorithms are measured in terms of the \textit{precision} and \textit{recall} criteria which are defined in terms of the inlier set ${{C_{in}}}$, the correctly judged set ${{C_{cor}}}$, and the ground truth set ${{C_{GT}}}$ as
\begin{IEEEeqnarray}{C}
{\rm{Precision}} = \frac{{\left| {{C_{cor}}} \right|}}{{\left| {{C_{in}}} \right|}},
\end{IEEEeqnarray}
and
\begin{IEEEeqnarray}{C}
{\rm{Recall}} = \frac{{\left| {{C_{cor}}} \right|}}{{\left| {{C_{GT}}} \right|}}.
\end{IEEEeqnarray}
\subsection{Dataset and experimental setup}
\subsubsection{Plant dataset}
\indent The models for this dataset have been obtained from the dataset containing 3D scans of plant shoot architectures provided by Salk Institute of Biological Science \cite{navlakha_2017}. This dataset contains $559$ 3D plant shoot architectures from 4 species (Arabidopsis, tomato, tobacco, and sorghum) scanned under multiple conditions (ambient light (control), heat, high-light, shade, drought). Each plant was scanned every 1 or 2 days through roughly 20 days of development.\\ In our work we have used 26 model of sorghum, tobacco and tomato plants in three different  conditions of control, shade and heat in different growth stage and different variety  because this data has the same categories and the same number of replicates which makes it easier to compare the results. The scenes are created by rotating the models and by using 
different levels of noise and down-sampling on these models, giving a total of $80$ scenes for each model under consideration.

\subsubsection{B3R dataset}
The Bologna 3D retrieval (B3R)
dataset \cite{tombari2013performance}, with 6 models, is used to evaluation the algorithms robustness with respect to noise and point densities. The models are taken from the Stanford Repository, however the scenes here are created by rotating the models, and then adding $10$ levels of noise and $8$ levels of downsamplig, giving a total of $80$ scenes for each model.
\begin{table}[ht!]
\caption{ Implementation platforms of the used techniques}
\label{table1}
\begin{center}
\begin{tabular}{l l}
\hline
Technique& \quad \quad Platform\\%
\hline
ISS Keypoints Detector& \quad \quad PCL\\
\hline
3D SIFT Descriptor& \quad \quad MATLAB\\%
\hline
NNSR& \quad \quad MATLAB\\%
\hline
RANSAC& \quad \quad MATLAB\\%
\hline
GC& \quad \quad MATLAB\\%
\hline
3D HV& \quad \quad MATLAB\\%
\hline
3D MLESAC& \quad \quad MATLAB\\%
\hline
\end{tabular}
\end{center}
\end{table}

\begin{table}[ht!]
\caption{Parameters of algorithms}
\label{table2}
\begin{center}
\begin{tabular}{l l}
\hline
NNSR& \quad \quad $t_{nnsr}$ \quad\quad \quad $20$ \quad \quad \\%
\hline
RANSAC& \quad \quad $t_{ransac}$ \quad \quad 0.01 \quad \quad $N_{ran}$ \quad \quad\quad 1000\\%
\hline
GC& \quad \quad $t_{GC}$ \quad\quad \quad 0.01\\

\hline
3D MLESAC& \quad \quad $t_{mlesac}$ \quad \quad 0.01 \quad \quad $N_{mlesac}$ \quad \quad 1000\\%
\hline
\end{tabular}
\end{center}
\end{table}
\begin{figure}[ht!]
    \centering
    \begin{subfigure}[b]{0.48\textwidth}
        \includegraphics[height=4.5cm, width=6.5cm]{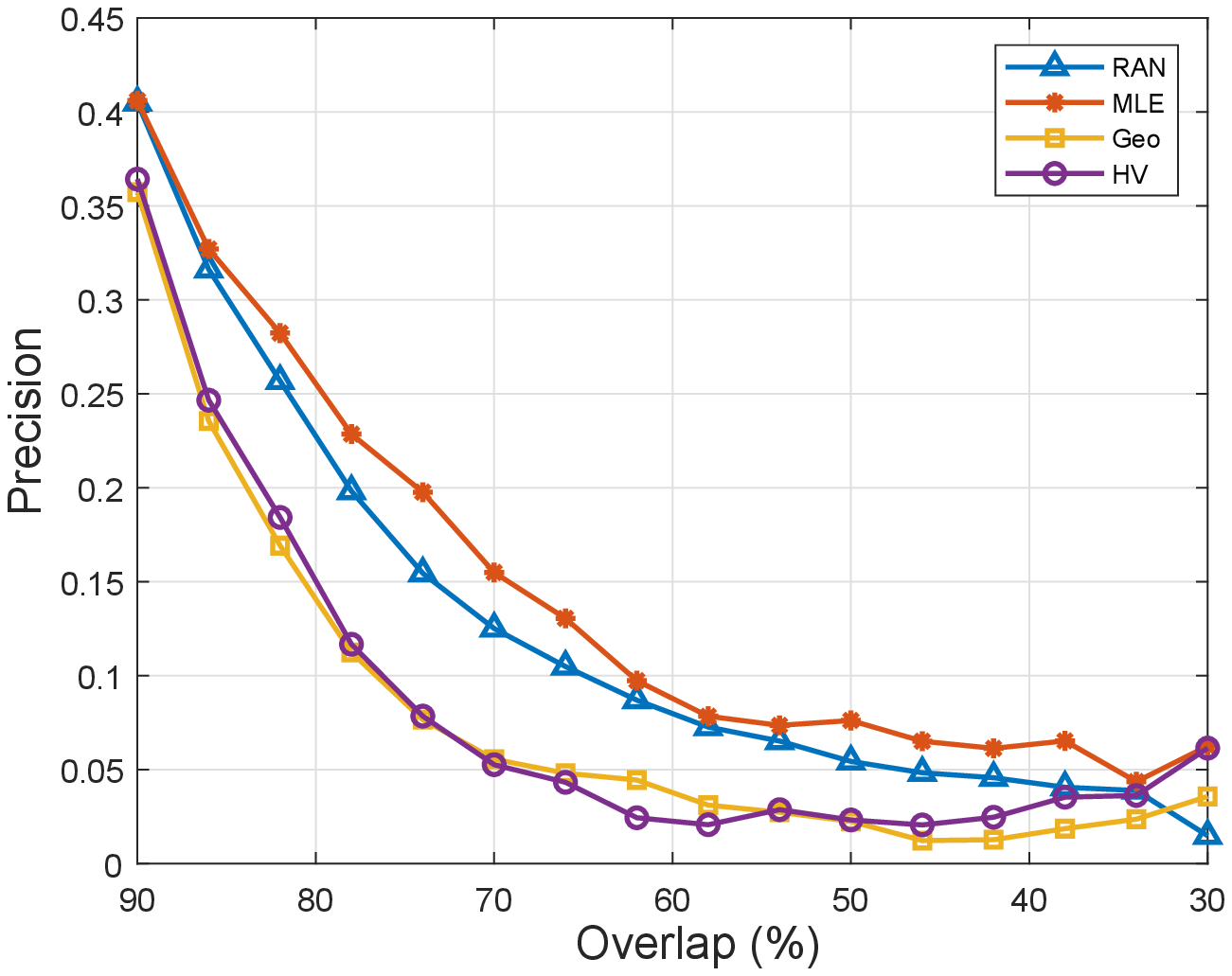}
        \label{fig:gull}
    \end{subfigure}
    ~ %
    \begin{subfigure}[b]{0.48\textwidth}
        \includegraphics[height=4.5cm, width=6.5cm]{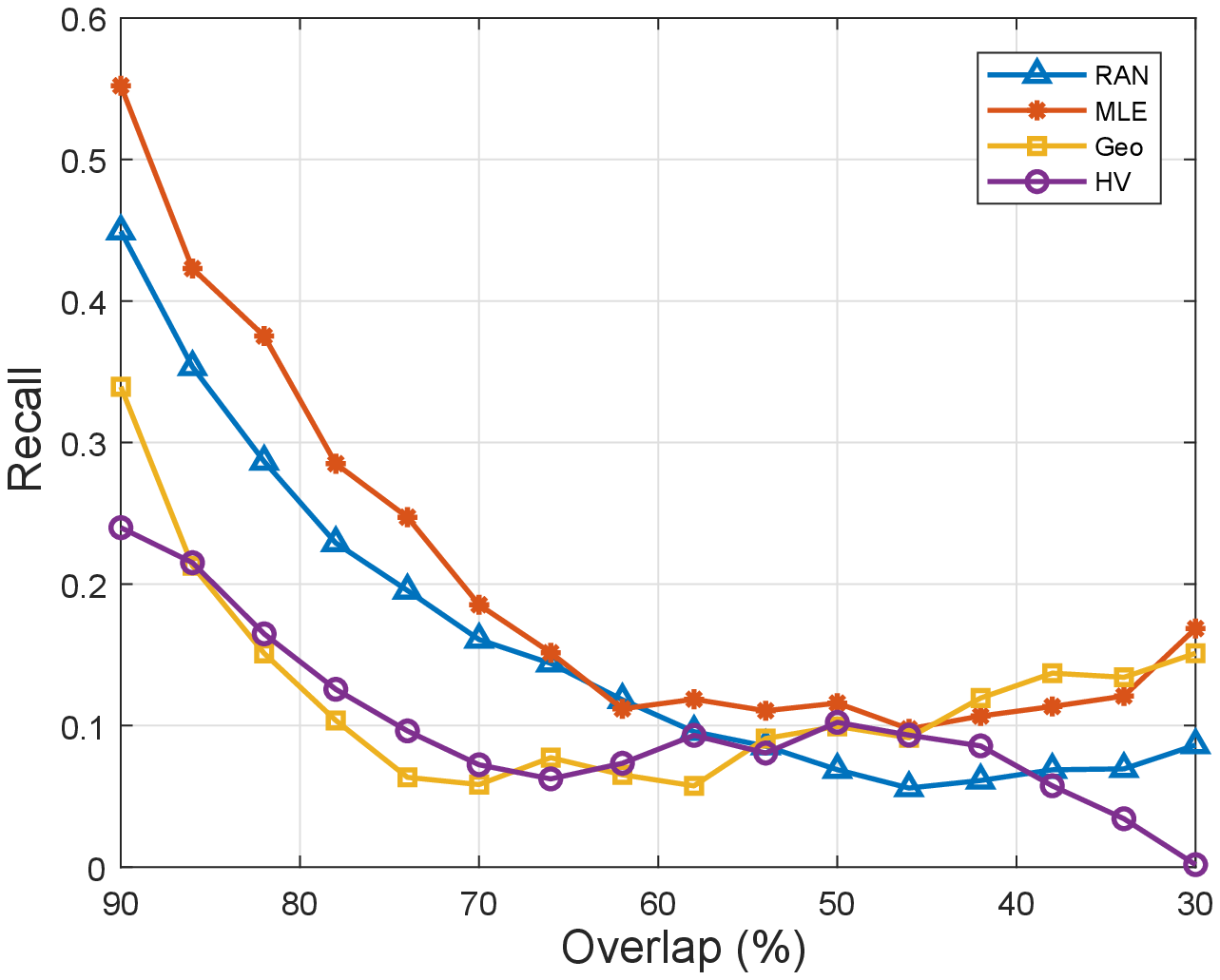}
        \label{fig:tiger}
    \end{subfigure}
    \caption{Performance of the evaluated CG algorithms in terms of
    precision and recall for the U3M dataset}\label{fig-u3m}
\end{figure}  
\begin{figure*}[!htbp]
    \centering
    \begin{subfigure}[b]{0.31\textwidth}
        \includegraphics[height=2.5cm, width=4.5cm]{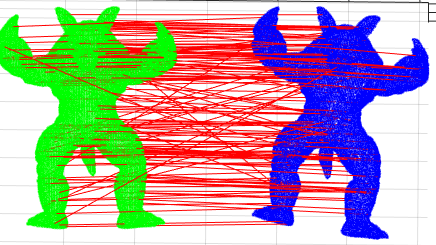}
        \caption{NNSR- B3R}
        \label{fig:gull}
    \end{subfigure}
    ~ %
    \begin{subfigure}[b]{0.31\textwidth}
        \includegraphics[height=2.5cm, width=4.5cm]{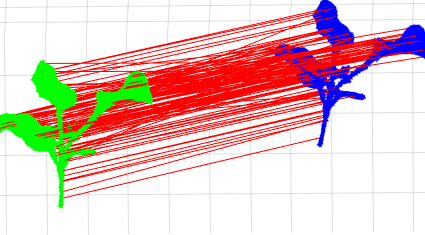}
        \caption{NNSR- Plant}
        \label{fig:tiger}
    \end{subfigure}
     \begin{subfigure}[b]{0.31\textwidth}
        \includegraphics[height=2.5cm, width=4.5cm]{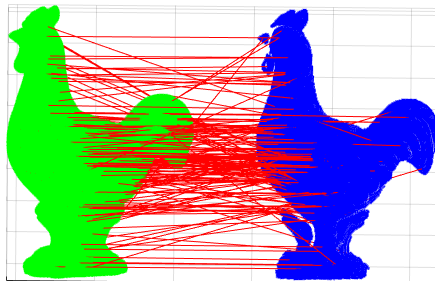}
       \caption{NNSR- U3M}
        \label{fig:tiger}
    \end{subfigure}
    \bigskip
    \begin{subfigure}[b]{0.31\textwidth}
        \includegraphics[height=2.5cm, width=4.5cm]{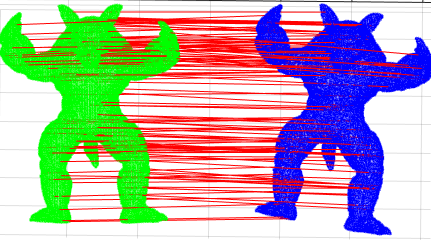}
       \caption{RANSAC- B3R}
        \label{fig:gull}
    \end{subfigure}
    ~ %
    \begin{subfigure}[b]{0.31\textwidth}
        \includegraphics[height=2.5cm, width=4.5cm]{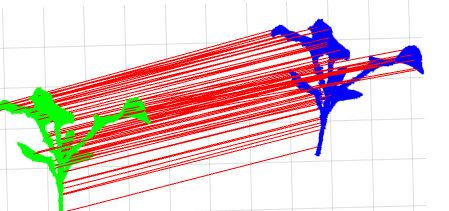}
       \caption{RANSAC- Plant}
        \label{fig:tiger}
    \end{subfigure}
     \begin{subfigure}[b]{0.31\textwidth}
        \includegraphics[height=2.5cm, width=4.5cm]{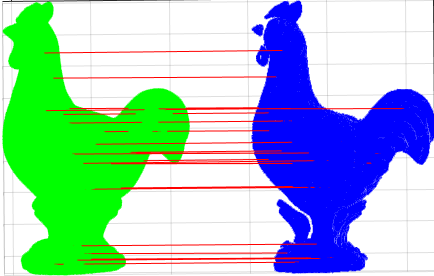}
        \caption{RANSAC- U3M}
        \label{fig:tiger}
    \end{subfigure}
    \bigskip
     \begin{subfigure}[b]{0.31\textwidth}
        \includegraphics[height=2.5cm, width=4.5cm]{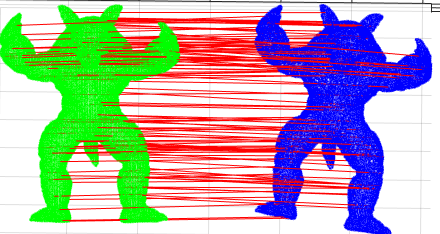}
       \caption{MLESAC- B3R}
        \label{fig:gull}
    \end{subfigure}
    ~ %
    \begin{subfigure}[b]{0.31\textwidth}
        \includegraphics[height=2.5cm, width=4.5cm]{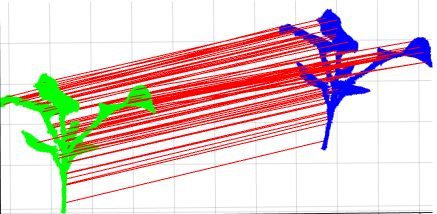}
       \caption{MLESAC- Plant}
        \label{fig:tiger}
    \end{subfigure}
     \begin{subfigure}[b]{0.31\textwidth}
        \includegraphics[height=2.5cm, width=4.5cm]{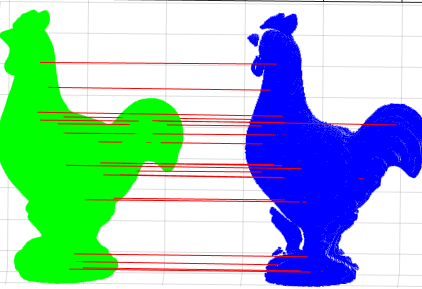}
        \caption{MLESAC- U3M}
        \label{fig:tiger}
    \end{subfigure}
    \bigskip
     \begin{subfigure}[b]{0.31\textwidth}
        \includegraphics[height=2.5cm, width=4.5cm]{ran_b3r.png}
        \caption{3DHV- B3R}
        \label{fig:gull}
    \end{subfigure}
    ~ %
    \begin{subfigure}[b]{0.31\textwidth}
        \includegraphics[height=2.5cm, width=4.5cm]{ran_plant.png}
        \caption{3DHV- Plant}
        \label{fig:tiger}
    \end{subfigure}
     \begin{subfigure}[b]{0.31\textwidth}
        \includegraphics[height=2.5cm, width=4.5cm]{ran_u3m.png}
        \caption{3DHV- U3M}
        \label{fig:tiger}
    \end{subfigure}
    \bigskip
     \begin{subfigure}[b]{0.31\textwidth}
        \includegraphics[height=2.5cm, width=4.5cm]{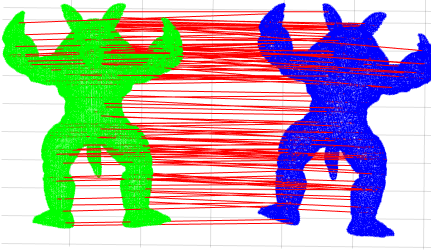}
       \caption{GC- B3R}
        \label{fig:gull}
    \end{subfigure}
    ~ %
    \begin{subfigure}[b]{0.31\textwidth}
        \includegraphics[height=2.5cm, width=4.5cm]{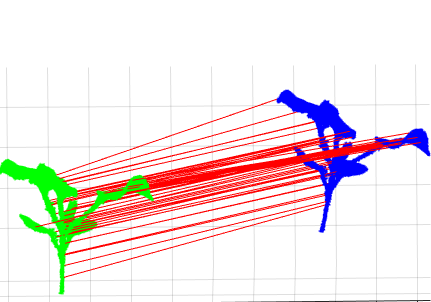}
        \caption{GC- Plant}
        \label{fig:tiger}
    \end{subfigure}
     \begin{subfigure}[b]{0.31\textwidth}
        \includegraphics[height=2.5cm, width=4.5cm]{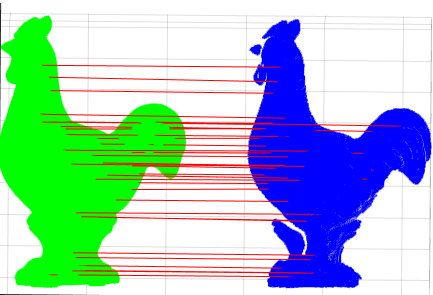}
        \caption{GC- U3M}
        \label{fig:tiger}
    \end{subfigure}
     \caption{Visualization of the evaluated CG algorithms for the three datasets}\label{fig-vis}
    \bigskip
\end{figure*} 
\subsubsection{U3M dataset}
The UWA 3D modeling (U3M) \cite{mian2006novel}
dataset belongs to the point cloud (2.5D view) registration scenario. There are $22$, $16$, $16$ and $21$ $2.5D$ views respectively captured from the Chef, Chicken, T-rex, and
Parasaurolophus models. There scenes represent different levels of overlaps. In our experiments, we have taken first $16$ views of each model.
\subsubsection{Implementations details}
In this paper, we have used 3D Intrinsic shape signature (ISS) for detecting keypoints and 3DSIFT for describing the local features. This choice is motivated by the results obtained in our previous work, where this detector-descriptor pair outperformed other combinations of popular detectors and descriptors in case of plant 3D point clouds. In the keypoint detection stage, around $3\%$ of points in the point cloud are chosen as interest points, with the non maximum radius 
set to round  $4$  point cloud resolution ($pr$). For the 3DSIFT descriptor, the search radius is set to $8pr$ and the size of feature descriptor is 128. 
NNSR is used for finding initial matching. RANSAC, GC, 3DHV, and 3D MLESAC are used as 3DCG algorithms. Table.\ref{table1} gives the implementation platforms of the used algorithms and the parameters used in all algorithms are listed in Table.\ref{table2}. 
\subsection{Results and discussions}
\subsubsection{Performance on the plant dataset}
The performance of the 3DCG algorithms for the plant dataset are shown in Fig.\ref{fig-plant}. Noise is expected to degrade on the discriminative capabilities of feature descriptor, thus
creating a certain amount of false matches. This is obvious in the plots obtained for precision and recall for all the 3DCG algorithms. The precision decreases with increasing noise variance. Similar trend is observed for the recall values, however, the deterioration is smaller, as compared to the precision. Down sampling the scenes also has negative impact on the performance, as seen in the precision and recall plots. This can be attributed to the false matches that become more common as the number of points become less. As indicated by the plots, RANSAC and MLESAC appear to be the best two ones among all evaluated algorithms,
considering their overall precision and recall performance, however, MLESAC generally outperforms RANSAC, while taking less time. This indicates that for larger point clouds of objects having highly complex shapes, MLESAC is a better choice as a 3DCG algorithm. 
\subsubsection{Performance on the B3R dataset}
The performance of the 3DCG algorithms for the B3R dataset are shown in Fig.\ref{fig-b3r}.
It follows similar trends, as in the case of the plant as seen in Fig.\ref{fig-plant}. As indicated by the plots  here also, RANSAC and MLESAC appear to be the best two ones among all evaluated algorithms, considering their overall precision and recall performance, with RANSAC performing slightly better. However, in our experiments, MLESAC is quite faster as compared to RANSAC.
\subsubsection{Performance on the U3M dataset}
A common trend to all algorithms is that their performance generally degrades as the degree of overlap drops as depicted in Fig.\ref{fig-u3m}. This is owing to the fact that the ratio of outliers in the initial correspondence set is closely correlated to the ratio of overlapping regions. MLESAC outperform in all algorithm in term of precious and recall.
\subsubsection{Visualization}
Fig.\ref{fig-vis} shows some visual results of 3DCG algorithms. These figures display the strength of each method. For example in NNSR method, The number of outliers are more than other methods, because of occlusion and overlap in the feature matching. We can also see the location and number of correspondences are also different.
\section{Conclusions}
In this paper, we
compared the performance of different 3D correspondence grouping algorithms in context of 3D plant point clouds and also provided the corresponding performance evolution of some standard publically available datasets for comparison. The performance was evaluated in terms of precision and recall, the results were presented in terms of the corresponding plots. Experimental results show that the RANSAC and MLESAC perform quite closely and are better than 3D Hough Voting and Geometric Consistency. Also, in general, MLESAC is much faster compared to RANSAC, making it a better algorithm for applications involving  3D point clouds of plants as they are usually very large in size. In future works, we intend to prepare our own dataset of 3D point clouds of crops such as rice and wheat and will try to extend the analysis further to perform plant identification for plant phenomics applications.

\bibliographystyle{unsrt}  
\bibliography{references}  

%
%
%
%

\end{document}